\documentclass[sigconf,screen]{acmart}
\AtBeginDocument{%
  }

\acmConference[MM '26]{The 34th ACM International Conference on Multimedia}{November 10--14,
  2026}{Rio de Janeiro, Brazil}
\setcopyright{none}
\renewcommand\footnotetextcopyrightpermission[1]{}
\begin{document}

\title{SinkRouter: Sink-Aware Routing for Efficient Long-Context Decoding in Large Language and Multimodal Models}
\author{Junnan Liu}
\orcid{0009-0009-2407-4403}
\email{25S30353@stu.hit.edu.cn}
\affiliation{%
  \institution{Harbin Institute of Technology (Weihai)}
  \city{Weihai}
  \country{China}
}

\author{Xinyan Liu}
\orcid{0000-0003-2638-4324}
\email{xinyliu@hit.edu.cn}
\affiliation{%
  \institution{Harbin Institute of Technology (Weihai)}
  \city{Weihai}
  \country{China}
}

\author{Peifeng Gao}
\email{gaopeifeng@connect.hku.hk}
\orcid{0009-0002-6231-4227}
\affiliation{%
  \institution{University of Hong Kong}
  \city{Hong Kong}
  \country{China}
}

\author{Zhaobo Qi}
\orcid{0000-0001-9196-9818}
\email{qizb@hit.edu.cn}
\affiliation{%
  \institution{Harbin Institute of Technology (Weihai)}
  \city{Weihai}
  \country{China}
}

\author{Beichen Zhang}
\orcid{0000-0001-5030-0632}
\email{beiczhang@hit.edu.cn}
\affiliation{%
  \institution{Harbin Institute of Technology (Weihai)}
  \city{Weihai}
  \country{China}
}

\author{Weigang Zhang}
\orcid{0000-0003-0042-7074}
\email{wgzhang@hit.edu.cn}
\affiliation{%
  \institution{Harbin Institute of Technology (Weihai)}
  \city{Weihai}
  \country{China}
}

\author{Antoni B. Chan}
\orcid{0000-0002-2886-2513}
\email{abchan@cityu.edu.hk}
\affiliation{%
  \institution{City University of Hong Kong}
  \city{Hong Kong}
  \country{China}
}

\begin{abstract}
In long-context decoding for LLMs and LMMs, attention becomes increasingly memory-bound because each decoding step must load a large amount of KV-cache data from GPU memory. Existing acceleration strategies often trade efficiency for accuracy by relying on heuristic pruning that may discard useful information. At a deeper level, they also tend to indiscriminately preserve all high-scoring tokens, treat early tokens as indispensable anchors, or rely on heuristic head routing, reflecting an insufficient mechanistic understanding of the attention sink phenomenon. In this paper, we show that the attention sink phenomenon corresponds to a stable, reachable, and error-controllable fixed point constructed during training. Based on this insight, we propose \textbf{SinkRouter}, a training-free selective routing framework that detects the sink signal and skips computations that would otherwise produce near-zero output. To translate this mechanism into real-world acceleration, we develop a \textbf{hardware-aware Triton kernel} with block-level branching and Split-K parallelism. We conduct extensive evaluations on a diverse suite of long-context benchmarks, including \textsc{LongBench}, \textsc{InfiniteBench}, \textsc{CVBench}, \textsc{MileBench}, and \textsc{MMVP}, using both text-only and multimodal backbones such as Llama-3.1-8B, Llama-3.1-70B, Yi-9B-200K, LLaVA-1.5-7B, and LLaVA-1.5-13B. Across these settings, SinkRouter consistently improves decoding efficiency while maintaining competitive accuracy, and reaches \textbf{2.03$\times$ speedup} with a \textbf{512K} context.
\end{abstract}





\maketitle

\section{Introduction}

Large Language Models (LLMs) and Large Multimodal Models (LMMs) are increasingly expected to process extended contexts, ranging from long textual histories to high-resolution images and videos~\cite{Dubey2024TheL3, touvron2023llama2openfoundation, ding2024longropeextendingllmcontext, hu2024mplugdocowl15unifiedstructure}. However, long-context autoregressive decoding introduces a critical systems bottleneck: at each decoding step, the model must repeatedly load an ever-growing Key-Value (KV) cache for attention computation. As the context length increases, this repeated fetching of the KV cache makes decoding increasingly memory-bound, thereby creating substantial pressure on memory bandwidth and incurring significant latency overhead on modern hardware~\cite{fu2024challengesdeployinglongcontexttransformers, pan2024instinferinstorageattentionoffloading,behnam2025rocketkv}.

A large body of recent work aims to alleviate this bottleneck through token eviction or KV-cache compression~\cite{liu2023scissorhands, xiao2024efficient, jo2026fastkvdecouplingcontextreduction, li2024snapkvllmknowslooking}. These approaches are effective in reducing memory footprint and decoding cost, but they primarily operate at the token or cache level. Consequently, they may discard context that later becomes useful, and their token-level sparsity patterns do not naturally align with the execution granularity of Grouped-Query Attention (GQA), which makes it difficult to consistently translate algorithmic sparsity into hardware speedup~\cite{zadouri2025hardwareefficient}.

Attention sinks have been widely observed in long-context decoding and are often associated with stabilization or anchor preservation~\cite{xiao2024efficient,yoo2026natureattentionsinkshapes}. Here, we revisit this phenomenon from a mechanistic perspective. During decoding, initial tokens such as BOS in Llama models often attract a disproportionately large share of attention mass~\cite{xiao2024efficient}. Our analysis suggests that this behavior is not merely a byproduct of the architecture, but is instead related to an adaptive bypass pattern acquired during training. Empirically, we find that the norm of the BOS value vector remains close to zero across layers, effectively acting as a numerical vacuum, while the BOS key remains geometrically separated from semantic tokens, making it a stable and easily reachable target. This combination allows certain attention heads to route toward BOS and produce an approximately null update to the residual stream when additional contextual enrichment is unnecessary. From this perspective, attention sinks provide a natural signal for identifying potentially idle heads during decoding.

Motivated by this observation, we propose \textbf{SinkRouter}, a plug-and-play, training-free decoding framework that uses sink-aware signals to selectively bypass weakly contributing attention heads. Rather than evicting tokens or compressing cached states, SinkRouter operates at the head level, reducing unnecessary KV-cache loading while preserving the retained context. To translate this selective computation into practical end-to-end acceleration, we further develop a hardware-aware Triton kernel with block-level uniform branching and Split-K parallelism, thereby enabling efficient execution under sparse routing decisions.

We evaluate SinkRouter on a diverse suite of long-context benchmarks, including \textsc{LongBench}~\cite{bai-etal-2024-longbench}, \textsc{InfiniteBench}~\cite{zhang-etal-2024-bench}, \textsc{MMVP}~\cite{tong2024eyeswideshutexploring}, \textsc{CVBench}~\cite{zhu2026cvbenchbenchmarkingcrossvideosynergies} and \textsc{MileBench}~\cite{dingjie2024milebench}, across both text-only and multimodal backbones, including Llama-3.1-8B, Llama-3.1-70B, Yi-9B-200K, LLaVA-1.5-7B, and LLaVA-1.5-13B. Across these settings, SinkRouter improves decoding efficiency while maintaining competitive task performance, achieving up to \textbf{2.03$\times$ speedup} with a 512K context on an NVIDIA RTX PRO 6000 GPU. These results demonstrate that SinkRouter provides a practical and mechanistically grounded approach to long-context decoding that is complementary to existing KV-cache reduction methods.

\section{Related Work}

\textbf{KV-Cache Reduction and Token/Visual Pruning.}
A major line of work accelerates long-context decoding by reducing the amount of historical information retained in memory. Representative methods include \textsc{Scissorhands}~\cite{liu2023scissorhands}, which prunes tokens based on a persistence-of-importance assumption, and \textsc{H2O}~\cite{zhang2023ho}, which retains heavy hitters according to cumulative attention scores. Other approaches further compress or approximate discarded context. For example, \textsc{LESS}~\cite{10.5555/3692070.3692524} summarizes evicted content into a constant-sized low-rank cache, while \textsc{SnapKV}~\cite{10.5555/3737916.3738638} selects important tokens from local prompt windows for subsequent decoding. In multimodal settings, methods such as \textsc{LOOK-M}~\cite{wan-etal-2024-look} and \textsc{FastV}~\cite{fastv} reduce visual redundancy through pruning or cache filtering. These methods primarily reduce decoding cost through token retention, pruning, or KV-cache compression. SinkRouter targets a different form of redundancy by performing selective computation at the head level during decoding, and can therefore be naturally combined with existing KV-cache reduction strategies.

\textbf{Attention Sinks and Their Interpretations.}
Attention sinks, in which initial tokens absorb a disproportionately large share of attention mass, have been widely observed in autoregressive decoding. \textsc{StreamingLLM}~\cite{xiao2024efficient} is the first method to exploit attention sinks as stable anchors for streaming inference. Subsequent studies have examined this phenomenon from several perspectives, including the conditions under which it emerges during training and its relation to softmax normalization~\cite{gu2025when}, as well as its functional role as an internal register or a structured global representation~\cite{darcet2024vision, kang2025see}. Other efforts attempt to weaken or avoid sink-related behavior through modified attention formulations~\cite{qiu2025gated, bondarenko2023quantizable}. More broadly, many score-based KV-cache eviction methods still rely on token-level attention magnitude as a proxy for importance, and therefore tend to retain high-attention tokens indiscriminately, even when some of them have limited influence on the final decoding outcome. This observation suggests that the mechanism underlying attention sinks is not yet fully captured by existing heuristics. SinkRouter is related to this line of research, but it uses sink-related behavior as a decoding-time signal for selective computation, rather than focusing on preserving, suppressing, or reparameterizing sink tokens themselves.

\textbf{Adaptive Computation and Head-Level Sparsity.}
Our work is also related to adaptive computation in Transformers, where computation is selectively allocated across heads or attention patterns. \textsc{FastGen}~\cite{ge2024model} exploits heterogeneity across attention heads and assigns different cache-retention strategies accordingly. \textsc{MInference}~\cite{MInference} identifies head-specific sparse attention patterns and accelerates long-context prefilling with optimized sparse kernels. \textsc{DejaVu}~\cite{liu2023dejavucontextualsparsity} uses learned predictors to select input-dependent subsets of attention heads and MLP parameters during inference. SinkRouter shares the same broad goal of reducing redundant computation, but differs in both routing criterion and execution granularity: it uses sink-aware signals to guide training-free selective computation during decoding at the head level.

\textbf{Hardware-Aware Attention and Sparse Execution.}
Practical decoding speedup often depends not only on the sparsity pattern itself, but also on whether the execution is supported by dedicated implementations. Although \textsc{FlashAttention}~\cite{dao2023flashattention2fasterattentionbetter} is highly optimized for dense attention, many sparse or selective attention methods require customized kernels or system support to realize their theoretical savings in practice. Recent examples include \textsc{NSA}~\cite{DBLP:conf/acl/YuanGD0ZZXWW0WR25}, which couples sparse attention design with hardware-aligned optimizations and specialized kernels. Related efforts such as \textsc{ShadowKV}~\cite{DBLP:conf/icml/SunCB0Z0DCC25} combine selective attention or sparse KV access with dedicated implementations to improve practical long-context inference efficiency. SinkRouter follows the same practical perspective: beyond its sink-aware routing rule, it includes a hardware-aware Triton implementation that translates reduced KV loading into end-to-end decoding gains.

\section{Insight}
\label{sec:insight}
Our work is motivated by three observations, which we examine empirically using representative text-only and multimodal models, including Llama-3.1-8B, Yi-9B-200K, LLaVA-1.5-7B, and LLaVA-1.5-13B.

\textbf{From Attention Sinks to a Low-Impact Update Regime.}
Our key hypothesis is that sink-dominant attention corresponds to a \emph{low-impact update regime}. When the current query does not require substantial contextual enrichment, softmax normalization still forces the model to allocate attention mass somewhere. Rather than assigning this mass to semantically active tokens and perturbing the residual stream, the model may route it to a sink state whose contribution to the layer output is small. From this perspective, attention sinks may reflect a learned strategy for producing controlled near-no-op updates during decoding, rather than merely an artifact of attention allocation.

\textbf{Why BOS-Dominant Routing Produces Weak Updates.}
To formalize this intuition, consider the attention output of head $h$ at layer $\ell$ during decoding step $t$. Let $x_t^{(\ell)}$ denote the current hidden state, and define the head update as
\begin{equation}
u_{\ell,h}(x_t^{(\ell)}) = \sum_{i=1}^{t} \alpha_i(x_t^{(\ell)})\, v_i^{(\ell,h)},
\label{eq:head_update}
\end{equation}
where the attention weights $\alpha_i$ are determined by the query--key similarities under the current hidden state. The corresponding residual update map is
\begin{equation}
F_{\ell,h}(x_t^{(\ell)}) = x_t^{(\ell)} + u_{\ell,h}(x_t^{(\ell)}).
\label{eq:fixed_point_map}
\end{equation}

\begin{figure}[tbp]
    \centering
    \includegraphics[width=\linewidth]{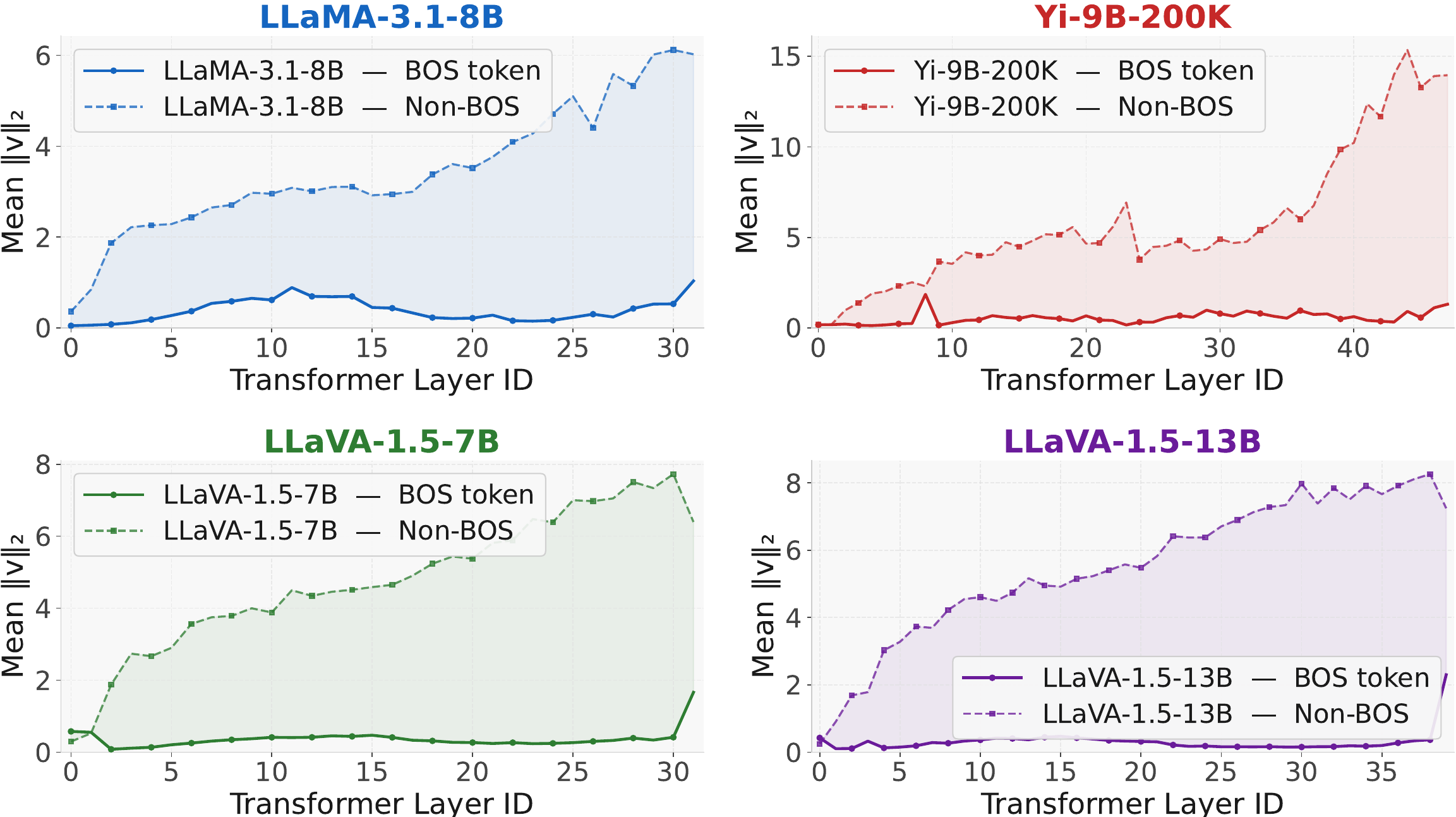}
    \caption{Mean $\|v\|_2$ across layers. BOS values remain close to zero, while non-BOS values have much larger norms.}
    %tokens exhibit much larger value norms.
    \Description{A line plot of mean value-vector norm across transformer layers. The BOS curve stays close to zero throughout all layers, while the curves for non-BOS tokens are consistently much higher, showing a clear separation between BOS and non-BOS value norms.}
    \label{fig:value_norm}
\end{figure}

A central empirical observation is that the BOS value vector has very small magnitude. As shown in Figure~\ref{fig:value_norm}, the norm $\|v_{\text{BOS}}\|_2$ remains close to zero across layers, whereas non-BOS value norms are substantially larger. Because the output of a head is a weighted sum of value vectors, routing most of the attention mass to BOS tends to produce a small residual update.

Table~\ref{tab:bos_attention_stats_llama31_8b} shows that when a head assigns a large attention score to BOS, the attention weights on non-BOS tokens become small. As a result, even if those tokens have large value magnitudes, their contributions to the final head output are suppressed by their low attention weights. Together with the near-zero magnitude of the BOS value vector, this attention pattern is likely to yield a small residual update. Consistent with this interpretation, results in Supplementary Section~B further show that heads with high BOS attention produce substantially smaller residual writes, and that their updates are also less aligned with the overall attention update direction of the layer. Taken together, these observations suggest that BOS-dominant heads tend to contribute less to the residual stream during decoding.

\begin{table}[t]
\centering
\caption{Attention concentration statistics on LLaMA-3.1-8B, measured from 500 LongBench samples. \textbf{Ratio} denotes the BOS attention score divided by the maximum non-BOS attention score.}
\label{tab:bos_attention_stats_llama31_8b}
\begin{tabular}{c|c|ccc}
\hline
Regime & Metric & 8K & 16K & 64K \\
\hline
BOS $\approx$ 0.4 & BOS   & 0.410 & 0.409 & 0.418 \\
                  & Max   & 2.90e-02 & 2.60e-02 & 3.65e-02 \\
                  & Mean  & 7.49e-05 & 3.76e-05 & 9.30e-06 \\
                  & Ratio & 16.4$\times$ & 17.8$\times$ & 12.9$\times$ \\
\hline
BOS $\approx$ 0.6 & BOS   & 0.598 & 0.594 & 0.600 \\
                  & Max   & 3.63e-02 & 3.81e-02 & 3.42e-02 \\
                  & Mean  & 5.12e-05 & 2.59e-05 & 6.39e-06 \\
                  & Ratio & 27.4$\times$ & 28.9$\times$ & 27.9$\times$ \\
\hline
BOS $\approx$ 0.8 & BOS   & 0.803 & 0.803 & 0.804 \\
                  & Max   & 1.82e-02 & 1.95e-02 & 1.89e-02 \\
                  & Mean  & 2.50e-05 & 1.26e-05 & 3.13e-06 \\
                  & Ratio & 51.8$\times$ & 51.0$\times$ & 50.5$\times$ \\
\hline
\end{tabular}
\end{table}

More formally, if
\begin{equation}
\alpha_{\text{BOS}}(x)\ge 1-\delta
\qquad\text{and}\qquad
\|v_{\text{BOS}}\|_2 \le \varepsilon_v,
\label{eq:bos_condition}
\end{equation}
then, by the triangle inequality,
\begin{equation}
\|u_{\ell,h}(x)\|_2
=
\Big\|
\alpha_{\text{BOS}}(x)v_{\text{BOS}}
+
\sum_{i\neq \text{BOS}} \alpha_i(x)v_i
\Big\|_2
\le
\varepsilon_v + \delta V_{\max},
\label{eq:update_bound}
\end{equation}
where
\begin{equation}
V_{\max}=\max_{i\neq \text{BOS}}\|v_i\|_2.
\label{eq:vmax}
\end{equation}
This bound supports an approximate fixed-point interpretation of sink-dominant routing. Since the residual update map is given by $F_{\ell,h}(x)=x+u_{\ell,h}(x)$, a small update norm implies that $F_{\ell,h}(x)$ remains close to $x$. Accordingly, when $\delta$ is small and $\|v_{\text{BOS}}\|_2$ remains close to zero, sink-dominant states may be viewed as $\varepsilon$-fixed points of the decode-time attention update, under which the residual stream is only weakly perturbed.

\begin{figure}[htbp]
    \centering
    \includegraphics[width=\linewidth]{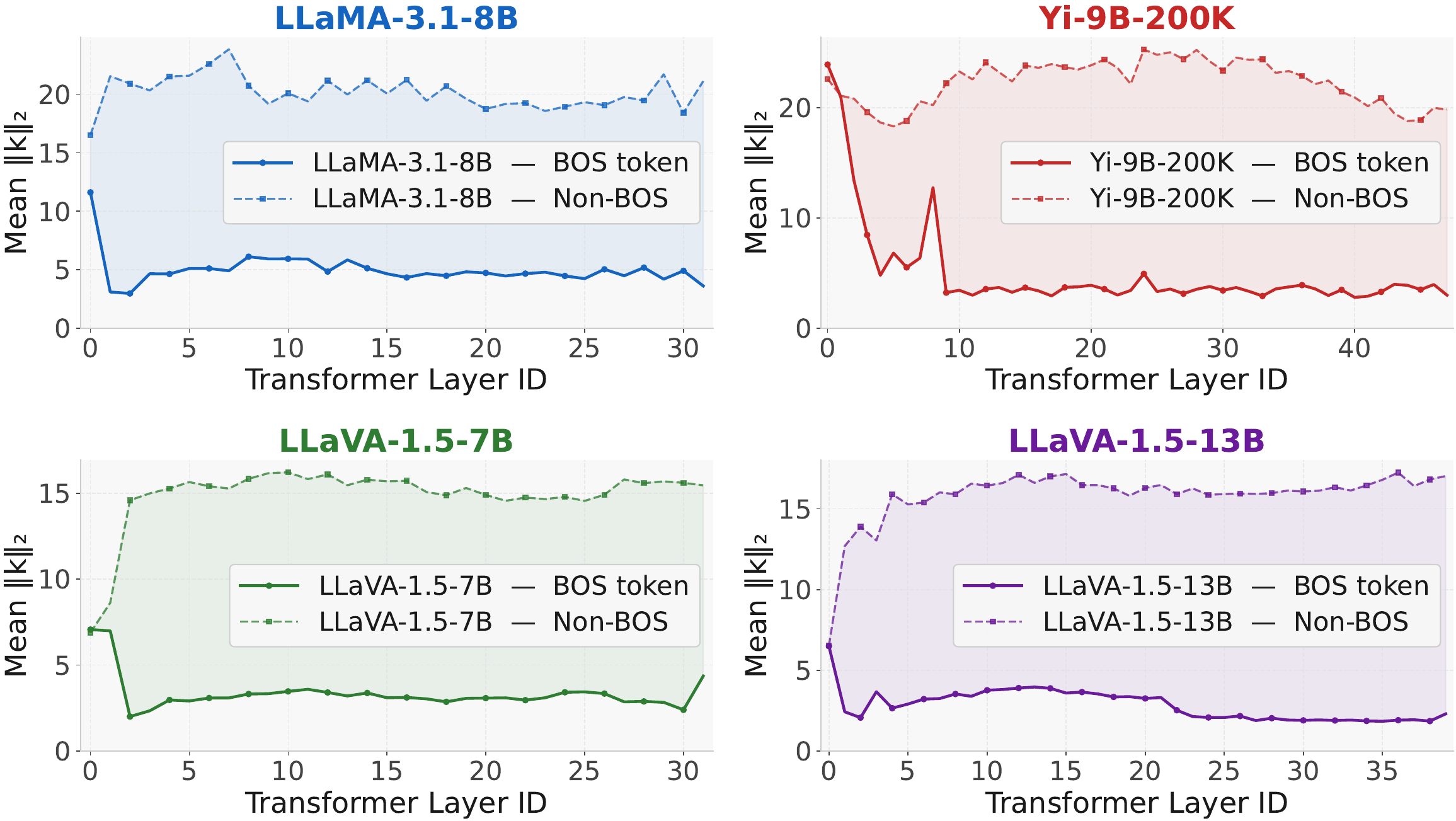}
    \caption{Mean $\|k\|_2$ across layers. BOS keys have smaller norms than semantic tokens, suggesting that sink dominance is not explained by key magnitude alone.}
    \Description{A line plot of mean key-vector norm across transformer layers for BOS and semantic tokens.}
    \label{fig:key_norm}
\end{figure}

\textbf{Why the Sink Is Reachable and Stable.}
A remaining question is why BOS can consistently receive such large attention mass. 
%We do not separately analyze $q_{\text{BOS}}$, because 
The sink behavior of %later 
the
decoding steps is governed by the interaction between the current query $q_t$ and the cached BOS key/value $(k_{\text{BOS}}, v_{\text{BOS}})$.
%, rather than by the BOS token's own query. 
The attention score between a query and a key is jointly determined by vector norms and directional alignment:
\begin{equation}
qk^\top = \|q\|_2 \|k\|_2 \cos(q,k).
\label{eq:dotproduct}
\end{equation}
If BOS were dominant because of an unusually large key norm, then the sink effect would be norm-driven. However, Figure~\ref{fig:key_norm} shows the opposite: the BOS key norm $\|k_{\text{BOS}}\|_2$ is smaller than that of semantic tokens. This suggests that key magnitude is unlikely to be the main explanation.

\begin{figure}[htbp]
    \centering
    \includegraphics[width=\linewidth]{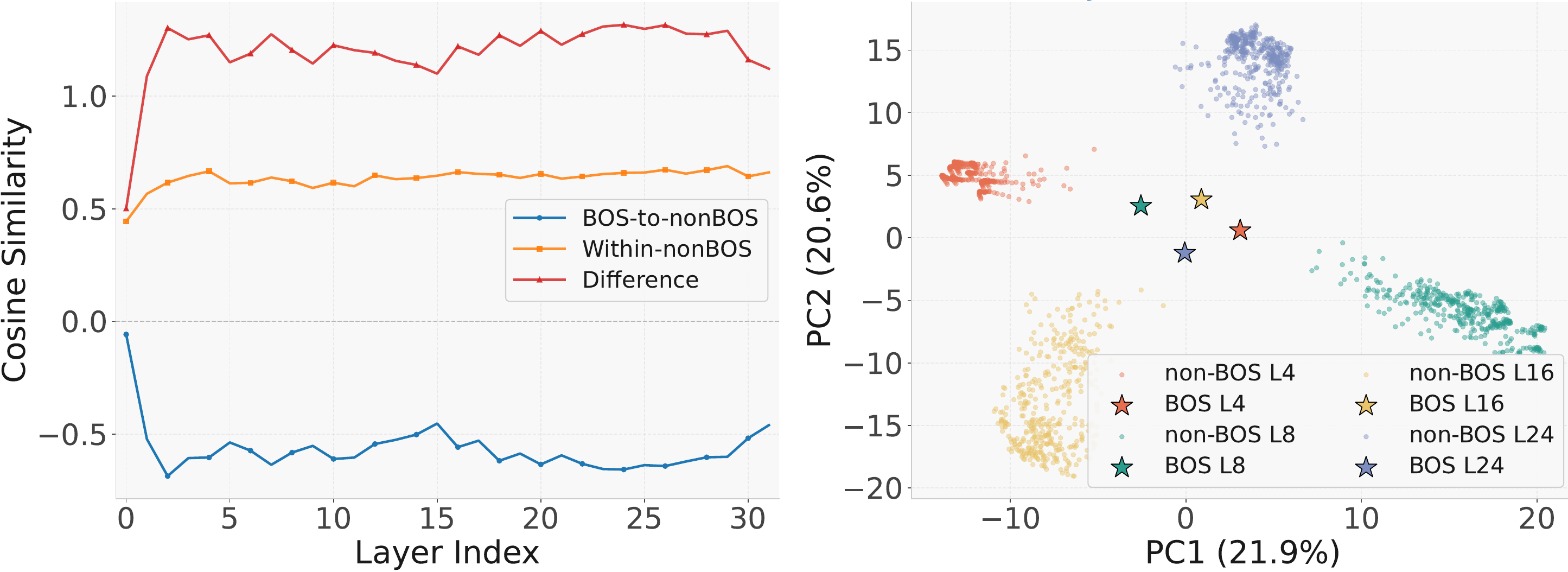}
    \caption{Cosine similarity and PCA of key vectors. BOS keys form a relatively isolated cluster away from the principal manifold of semantic tokens.}
    \Description{A visualization of BOS and semantic key vectors using cosine similarity analysis and principal component analysis.}
    \label{fig:pca_keys}
\end{figure}

The dominant factor is therefore more likely to be directional alignment. If BOS receives a large attention score despite its relatively small key norm, the query is likely to be aligned with it in direction.
This interpretation is consistent with the geometry of the key space. Figure~\ref{fig:pca_keys} suggests that BOS keys form a compact and stable region in the key space. In the PCA projection, this region appears relatively concentrated and close to the global mean of keys, which is consistent with the hypothesis that BOS keys are largely shaped by a shared, bias-like component rather than token-specific semantic content. Such geometric isolation makes BOS a plausible routing target when a token requires limited contextual retrieval, because the query can move toward this separated BOS direction without directly competing with the dense semantic-token manifold. Additional threshold calibration and stability results in Supplementary Section~A further indicate that this sink-oriented routing behavior recurs across inputs and can be regulated in a relatively predictable manner under a fixed threshold.

Taken together, these observations suggest that attention sinks may be understood as geometrically isolated, low-impact $\varepsilon$-fixed points to which the Transformer can route when additional contextual computation appears unnecessary.

% As further analyzed in the supplementary material Section B, query-head regions with higher BOS dominance exhibit smaller surrogate-induced output perturbation.

\section{Method}
\label{sec:method}

Based on the mechanistic perspective developed in Section~\ref{sec:insight}, we propose \textbf{SinkRouter}, a training-free, plug-and-play selective routing framework for long-context decoding. Before loading the historical KV cache, SinkRouter estimates whether the current KV group is likely to enter a sink-dominant, low-impact regime. If so, it skips historical KV-cache loading and replaces the corresponding attention output with a cheap surrogate; otherwise, it falls back to standard attention. In this way, SinkRouter uses sink-related behavior as an execution-time signal for decoding acceleration.

\begin{figure*}[htbp]
\centering
\includegraphics[width=\linewidth]{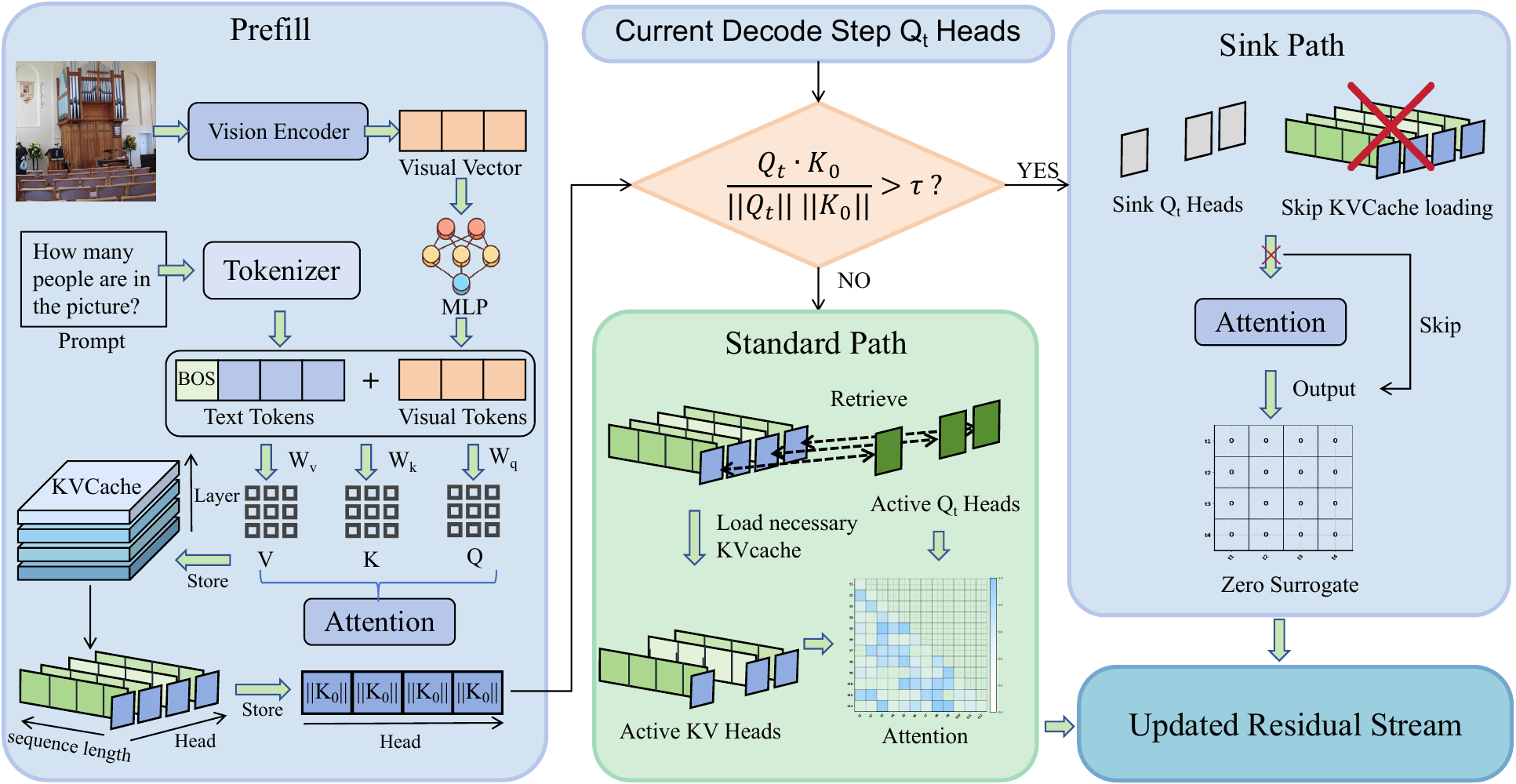}
\caption{Workflow of SinkRouter. During prefill, the model performs standard full attention and extracts lightweight initial-token anchor metadata for each KV group. During decoding, SinkRouter uses these anchors to identify sink-oriented groups before historical KV-cache loading. Sink groups exit early with a zero surrogate, while active groups proceed with standard attention.}
\Description{A workflow diagram of SinkRouter with two stages: prefill and decode. In the prefill stage, the model runs standard full attention and extracts lightweight anchors associated with the first cached token for each KV group. In the decode stage, these anchors are used before loading the historical KV cache to identify sink-oriented groups. Sink groups exit early and use a zero surrogate, while active groups continue with standard attention computation. The figure highlights that SinkRouter reduces unnecessary KV-cache loading by making routing decisions before historical attention is executed.}
\label{fig:overview}
\end{figure*}

A central design principle of SinkRouter is that all routing decisions are made at the \emph{KV-group level}. Under grouped-query attention (GQA), multiple query heads share the same KV cache, making the KV group the natural unit of memory access and execution. SinkRouter therefore performs routing, calibration, and evaluation consistently at the group level rather than at the individual-head level. The overall workflow is illustrated in Figure~\ref{fig:overview}.

\subsection{Prefill Phase: Initial-Token Anchor Caching}

For notational generality, we use token $0$ to denote the first cached token in each KV group. In the models studied in this work, this token coincides with BOS. SinkRouter leaves the prefill phase unchanged so that prompt processing remains identical to standard full attention. The only additional step is to cache lightweight metadata associated with token $0$ in each layer $l$ and KV group $g$:
\begin{equation}
\|K_0^{(l,g)}\|_2
\end{equation}
Since this metadata corresponds to only one token per KV group, its memory overhead is negligible relative to the full KV cache.

These anchors provide the minimal information needed for decode-time routing. In the models studied here, Section~\ref{sec:insight} shows that sink-dominant behavior is associated with strong alignment to the BOS direction and a low-impact attention output. SinkRouter uses the cached initial-token anchors to detect this regime cheaply during decoding, without scanning the full historical context.

\subsection{Decode-Time Group Routing}

During autoregressive decoding, the dominant cost comes from repeatedly attending over the growing KV cache. SinkRouter reduces this cost by determining, for each KV group, whether its historical KV cache needs to be loaded.

Let $H_q$ and $H_{kv}$ denote the numbers of query heads and KV heads, respectively, and let
\begin{equation}
r = \frac{H_q}{H_{kv}}
\end{equation}
be the number of query heads associated with each KV group. For group $g$, we denote the corresponding set of query heads by
\begin{equation}
\mathcal{Q}_g = \{q_{g,1}, q_{g,2}, \dots, q_{g,r}\}.
\end{equation}

\paragraph{Target regime.}
We first define the target regime that SinkRouter aims to detect. For a query head $q$, let $\alpha_0$ denote the attention mass assigned to token $0$ under full attention. We regard the query as sink-like if
\begin{equation}
\alpha_0 > \gamma,
\label{eq:oracle_sink}
\end{equation}
where $\gamma$ is an initial-token dominance threshold.

This corresponds to the regime identified in Section~\ref{sec:insight}: once token $0$ receives sufficiently large attention mass, the non-initial-token attention weights become small and the resulting attention output is expected to have low impact. In the models studied in this work, token $0$ coincides with BOS, and Table~\ref{tab:bos_attention_stats_llama31_8b} shows that around $\alpha_{\text{BOS}}\approx 0.6$, the competing non-BOS attention weights are already very small. We therefore use a slightly more conservative value, $\gamma=0.65$, as the operational threshold.

\paragraph{Routing signal.}
The oracle event in Eq.~\ref{eq:oracle_sink} cannot be evaluated before attention is computed, so SinkRouter uses a lightweight proxy based on alignment with the initial-token anchor. For each query head $q_{g,i}\in\mathcal{Q}_g$, we compute
\begin{equation}
s_{g,i} = \cos(q_{g,i}, K_0^{(l,g)})
= \frac{q_{g,i} K_0^{(l,g)\top}}
{\|q_{g,i}\|_2 \|K_0^{(l,g)}\|_2}.
\label{eq:cos_routing}
\end{equation}
This choice follows naturally from Section~\ref{sec:insight}: in the models studied here, the initial-token key does not have an unusually large norm, so sink-dominant behavior is more naturally associated with directional alignment than with key magnitude. The score in Eq.~\ref{eq:cos_routing} is inexpensive to compute and avoids any scan over the historical KV cache.

\paragraph{Group-level decision.}
Because all query heads in $\mathcal{Q}_g$ share the same KV cache, the routing decision must be made jointly at the KV-group level. We therefore aggregate the query-level scores into a single group score:
\begin{equation}
S_g = \frac{1}{r}\sum_{q_{g,i}\in\mathcal{Q}_g} s_{g,i}.
\label{eq:group_score}
\end{equation}
We then classify group $g$ as a \textbf{Sink Group} if
\begin{equation}
S_g > \tau(L),
\label{eq:sink_group}
\end{equation}
and as an \textbf{Active Group} otherwise, where $\tau(L)$ is a length-dependent routing threshold.

\paragraph{Threshold calibration.}
The score in Eq.~\ref{eq:cos_routing} is only a proxy, so the threshold $\tau(L)$ must be calibrated. A single global threshold is insufficient because the operating point changes systematically with context length. Supplementary Section~A shows that the threshold required to realize a fixed skip budget varies systematically (and non-monotonically) with context length while exhibiting only limited variation across tasks. We therefore calibrate a length-adaptive threshold $\tau(L)$ once per model as a function of context length $L$ by determining the threshold required to realize the target skip ratio at a small set of representative lengths and fitting the resulting calibration points with a cubic function of normalized context length.

\paragraph{Execution policy.}
For Sink Groups, SinkRouter skips historical KV-cache loading and replaces the corresponding attention output with a zero surrogate:
\begin{equation}
\mathrm{Output}(q_{g,i}) \approx \mathbf{0},
\qquad \forall q_{g,i}\in\mathcal{Q}_g \text{ if } S_g > \tau(L).
\label{eq:zero_approx}
\end{equation}
This surrogate is motivated by the low-impact regime characterized in Section~\ref{sec:insight}. In the models studied here, token $0$ coincides with BOS, whose value vector exhibits near-zero magnitude under sink-dominant routing. The zero surrogate therefore serves as a simple and efficient approximation in this regime.

For Active Groups, SinkRouter falls back to standard attention over the available KV cache. In this way, SinkRouter preserves exact computation where contextual retrieval is still likely to matter, while turning sink-dominant groups into a cheap decode-time bypass.

\subsection{System-Level Optimization}

Group-level sparsity alone does not guarantee wall-clock speedup. In decode-time attention, the critical path is governed largely by repeated KV-cache movement rather than arithmetic alone. To translate sink-aware sparsity into practical acceleration, we implement SinkRouter as a Triton kernel that treats the KV group as the basic execution unit and performs routing before historical KV-cache loading.

\paragraph{Group-centric execution unit.}
Our implementation follows the native structure of GQA. Queries are organized as
\[
Q \in \mathbb{R}^{B \times H_{kv} \times G \times D},
\]
where $G$ is the number of query heads that share one KV head, while the cached keys and values are stored as
\[
K, V \in \mathbb{R}^{B \times H_{kv} \times L \times D},
\]
and the initial-token anchor is stored as
\[
K_0 \in \mathbb{R}^{B \times H_{kv} \times D}.
\]
This layout keeps the shared-KV structure explicit throughout execution and avoids explicit \texttt{repeat\_kv}-style expansion.

\paragraph{Inline routing before historical KV loading.}
Routing is evaluated inside the attention kernel, before any historical KV-cache block is fetched. For each program instance, the kernel first loads the initial-token key of the current KV group, evaluates the alignment score for the query heads in that group, and applies the routing condition. If the group is classified as a Sink Group, the kernel directly writes the zero surrogate and returns without loading historical $K$ or $V$. This removes the need for a separate routing kernel and ensures that skipped groups avoid historical KV-cache transfer altogether.
\begin{table*}[t]
\centering
\small
\caption{Detailed results on LongBench. Scores are reported in percentage. $\Delta$ Avg. denotes the difference from the corresponding Full Attention baseline within each model block.}
\label{tab:longbench_detailed}
\begin{tabular}{lcccccccccc}
\toprule
\textbf{Method} & \textbf{NQA} & \textbf{MF-en} & \textbf{HP} & \textbf{MQ} & \textbf{DR} & \textbf{GR} & \textbf{PR} & \textbf{LCC} & \textbf{Avg.} & \textbf{$\Delta$ Avg.} \\
\midrule
\multicolumn{11}{l}{\textbf{Llama-3.1-8B}} \\
Full Attention & 30.00 & 55.14 & 56.15 & 30.08 & 33.85 & 35.20 & 99.50 & 64.73 & 50.58 & -- \\
H2O            & 29.69 & 54.99 & 56.10 & 30.55 & 34.11 & 35.15 & 99.50 & 64.81 & 50.61 & +0.03$\uparrow$ \\
FastGen        & 26.02 & 52.92 & 52.79 & 26.54 & 18.75 & 31.61 & 92.50 & 64.21 & 45.67 & -4.91$\downarrow$ \\
StreamingLLM   & 20.70 & 38.05 & 40.18 & 15.90 & 14.93 & 31.29 & 36.50 & 65.03 & 32.82 & -17.76$\downarrow$ \\
SinkRouter(Ours)           & 30.08 & 53.96 & 55.71 & 29.62 & 32.14 & 33.87 & 99.50 & 64.24 & 49.89 & -0.69$\downarrow$ \\
\midrule
\multicolumn{11}{l}{\textbf{Yi-9B-200K}} \\
Full Attention & 11.92 & 37.60 & 51.66 & 27.55 & 21.66 & 30.62 & 59.50 & 73.62 & 39.27 & -- \\
H2O            & 12.25 & 36.56 & 50.73 & 26.95 & 21.04 & 29.30 & 58.00 & 72.84 & 38.46 & -0.81$\downarrow$ \\
FastGen        & 10.65 & 34.29 & 43.97 & 17.52 & 12.94 & 24.43 & 33.50 & 71.87 & 31.15 & -8.12$\downarrow$ \\
StreamingLLM   & 4.29  & 31.17 & 43.07 & 17.93 & 13.75 & 22.61 & 19.50 & 62.83 & 26.89 & -12.38$\downarrow$ \\
SinkRouter(Ours)           & 11.66 & 37.24 & 52.01 & 27.81 & 20.81 & 30.13 & 50.50 & 73.22 & 37.92 & -1.35$\downarrow$ \\
\midrule
\multicolumn{11}{l}{\textbf{Llama-3.1-70B}} \\
Full Attention & 34.81 & 54.24 & 64.45 & 47.08 & 32.70 & 34.93 & 98.00 & 70.88 & 54.64 & -- \\
H2O            & 34.92 & 54.31 & 64.45 & 47.08 & 32.53 & 34.92 & 98.00 & 69.73 & 54.49 & -0.15$\downarrow$ \\
FastGen        & 30.24 & 52.32 & 62.96 & 41.28 & 19.62 & 31.62 & 95.29 & 68.01 & 50.17 & -4.47$\downarrow$ \\
StreamingLLM   & 25.35 & 46.42 & 55.23 & 26.15 & 16.39 & 33.41 & 60.50 & 68.95 & 41.55 & -13.09$\downarrow$ \\
SinkRouter(Ours)           & 34.43 & 54.67 & 64.35 & 46.75 & 33.01 & 34.84 & 98.00 & 70.46 & 54.56 & -0.08$\downarrow$ \\
\bottomrule
\end{tabular}
\end{table*}

\paragraph{Split-K attention for active groups.}
If a group is classified as active, SinkRouter executes attention with Split-K parallelism along the sequence dimension. The sequence of length $L$ is partitioned into \texttt{num\_splits} chunks, and each program instance processes one chunk of the shared KV cache:
\[
[\texttt{chunk\_start}, \texttt{chunk\_end}) \subseteq [0, L).
\]
Within each split, the kernel loads one query tile $Q \in \mathbb{R}^{G \times D}$ and iterates over contiguous KV blocks $K,V \in \mathbb{R}^{B_K \times D}$ using online softmax accumulation. This preserves contiguous KV fetching on the active path and helps maintain hardware utilization when many sink groups exit early.

Taken together, these system-level optimizations align SinkRouter with the shared-KV structure of GQA, reduce unnecessary KV-cache traffic, and help translate sink-aware sparsity into practical decoding speedup.

\section{Experiments}
\subsection{Experimental Setup}
\textbf{Models.} We evaluate SinkRouter on both text-only and multimodal backbones. For text-only evaluation, we use Llama-3.1-8B, Llama-3.1-70B, and Yi-9B-200K. For multimodal evaluation, we use LLaVA-1.5-7B and LLaVA-1.5-13B. All models are evaluated using their pretrained checkpoints without finetuning, and SinkRouter is applied only at inference time.

\textbf{Benchmarks.} We evaluate text-only models on \textsc{LongBench}~\cite{bai-etal-2024-longbench} and \textsc{InfiniteBench}~\cite{zhang-etal-2024-bench}. For multimodal models, we use \textsc{MMVP}~\cite{tong2024eyeswideshutexploring}, \textsc{CVBench}~\cite{zhu2026cvbenchbenchmarkingcrossvideosynergies}, and \textsc{MileBench}~\cite{dingjie2024milebench}. \textsc{LongBench} is used to assess general long-context task performance, while \textsc{InfiniteBench} is used to probe more demanding long-context retrieval and precision-sensitive settings. For multimodal models, \textsc{MileBench}, \textsc{CVBench} and \textsc{MMVP} are used to test whether the same decode-time routing mechanism transfers to multimodal backbones. We follow the official evaluation protocols and metrics of each benchmark.

\textbf{Baselines.} We compare SinkRouter with the Full Attention baseline and representative training-free decoding baselines. For text-only evaluation, we include \textsc{FastGen}~\cite{ge2024model}, \textsc{StreamingLLM}~\cite{xiao2024efficient} and  \textsc{H2O}~\cite{zhang2023ho}. For multimodal evaluation, we include \textsc{LOOK-M}~\cite{wan-etal-2024-look} and \textsc{FastV}~\cite{fastv}. All baselines are run using their official implementations.

\textbf{Budget alignment and fairness.} For fair comparison, we align the \emph{effective KV-cache budget} across methods rather than their method-specific hyperparameters. In our setup, SinkRouter retains about 40\% of heads/groups during decoding, corresponding to an effective KV-cache budget of approximately 40\%. We therefore configure the baselines to match this budget whenever applicable. Specifically, for \textsc{H2O}, we allocate 20\% of the cache to recent tokens and 20\% to heavy hitters; for \textsc{StreamingLLM}, we retain the first 4 sink tokens and set the recent window to 40\% of the cache; and for the multimodal baselines \textsc{LOOK-M} and \textsc{FastV}, we use a 40\% compression budget. SinkRouter is calibrated with a routing threshold that realizes this retained-head budget in practice. Although the resulting sparsity patterns differ across methods, all methods are evaluated under closely matched resource constraints.

\textbf{Implementation and hardware.} All efficiency results are measured on an NVIDIA RTX PRO 6000 GPU. All experiments are run in bf16 precision. 

\subsection{Accuracy Evaluation}

We evaluate whether SinkRouter preserves task performance on both text-only and multimodal benchmarks under matched decoding budgets. Overall, SinkRouter remains close to the corresponding Full Attention baselines across most settings, indicating that sink-aware selective computation introduces only limited accuracy degradation in practice.

Table~\ref{tab:longbench_detailed} reports the text-only results on LongBench. Across all three backbones, SinkRouter remains close to the corresponding Full Attention baseline on the benchmark average. This pattern becomes clearer when compared with more aggressive approximation methods: FastGen and StreamingLLM show substantially larger drops, especially on the smaller and mid-sized models, whereas SinkRouter stays within a much narrower range of the original model performance. H2O is also generally competitive with Full Attention, but SinkRouter achieves similarly stable behavior through a different mechanism. Overall, these results suggest that SinkRouter preserves long-context task performance with smaller average degradation than stronger token-eviction baselines.

\begin{table}[t]
\centering
\small
\caption{Results on InfiniteBench for LLaMA-3.1-8B. Scores are reported in percentage. $\Delta$ Avg. is computed relative to the Full Attention baseline.}
\label{tab:infinitebench_llama31_8b}
\begin{tabular}{lccc}
\toprule
\textbf{Dataset} & \textbf{Full Attention} & \textbf{FastGen} & \textbf{SinkRouter(Ours)} \\
\midrule
passkey               & 100.00 & 99.00 & 100.00 \\
number\_string        & 99.49  & 99.00 & 99.49 \\
kv\_retrieval         & 54.60  & 40.50 & 56.80 \\
code\_debug           & 23.52  & 21.50 & 22.87 \\
math\_find            & 49.50  & 34.50 & 50.00 \\
longdialogue\_qa\_eng & 19.00  & 17.00 & 19.00 \\
longbook\_choice\_eng & 68.50  & 69.00 & 68.00 \\
longbook\_qa\_eng     & 28.52  & 27.20 & 27.43 \\
longbook\_qa\_chn     & 37.74  & 37.66 & 37.66 \\
\midrule
\textbf{Avg.}          & \textbf{53.43} & \textbf{49.48} & \textbf{53.47} \\
\textbf{$\Delta$ Avg.} & \textbf{--} & \textbf{-3.95$\downarrow$} & \textbf{+0.04$\uparrow$} \\
\bottomrule
\end{tabular}
\end{table}

Table~\ref{tab:infinitebench_llama31_8b} shows a similar trend on InfiniteBench. SinkRouter remains almost identical to Full Attention on the benchmark average and stays competitive on precision-sensitive retrieval-style tasks. At the same time, the results do not indicate a uniform improvement over Full Attention on every task. A more accurate interpretation is that SinkRouter introduces very limited disturbance to the original retrieval behavior while remaining more robust than FastGen in these extreme-context settings.

\begin{table}[t]
\centering
\small
\caption{Results on MMVP and CV-Bench. Scores are reported in percentage. $\Delta$ Avg. denotes the difference from the corresponding Full Attention baseline within each model block.}
\label{tab:mmvp}
\begin{tabular}{lccccc}
\toprule
\textbf{Method} & \textbf{MMVP} & \textbf{CV-Bench$^{2D}$} & \textbf{CV-Bench$^{3D}$} & \textbf{Avg.} \\
\midrule
\multicolumn{5}{l}{\textbf{LLaVA-1.5-7B}} \\
Full Attention & 53.44 & 46.87 & 61.83 & 54.05  \\
SinkRouter(Ours)           & 52.63 & 47.15 & 61.67 & 53.82  \\
\midrule
\multicolumn{5}{l}{\textbf{LLaVA-1.5-13B}} \\
Full Attention & 60.73 & 59.25 & 62.08 & 60.69  \\
SinkRouter(Ours)           & 60.73 & 58.97 & 61.92 & 60.54  \\
\bottomrule
\end{tabular}
\end{table}

Table~\ref{tab:mmvp} reports the basic multimodal perception results on MMVP and CV-Bench. SinkRouter remains very close to the corresponding LLaVA baselines on both the 7B and 13B models. The average differences are small, and there is no sign of systematic degradation in either the 2D or 3D subsets. This suggests that the routing mechanism primarily affects redundant computation during decoding, while leaving the underlying visual-language alignment largely unchanged.

\begin{table*}[t]
\centering
\caption{Results on MileBench. Scores are reported in percentage. $\Delta$ Avg. denotes the difference from the corresponding Full Attention baseline within each model block.}
\label{tab:llava-combined-transposed}
\resizebox{\textwidth}{!}{
\begin{tabular}{lccccccccc}
\toprule
\textbf{Method} & \textbf{ObjectExistence} & \textbf{ActionLocalization} & \textbf{DocVQA} & \textbf{TQA} & \textbf{SpotTheDiff} & \textbf{CLEVR-Change} & \textbf{MMCoQA} & \textbf{Avg.} & \textbf{$\Delta$ Avg.} \\
\midrule
\multicolumn{10}{l}{\textbf{LLaVA-1.5-7B}} \\
Full Attention & 51.5 & 24.0 & 43.5 & 42.0 & 15.8 & 14.7 & 40.5 & 33.14 & -- \\
FastV          & 51.0 & 26.5 & 43.0 & 41.5 & 16.2 & 15.3 & 41.5 & 33.57 & +0.43$\uparrow$ \\
LOOK-M         & 50.5 & 24.0 & 41.0 & 38.5 & 17.1 & 11.9 & 9.5  & 27.50 & -5.64$\downarrow$ \\
SinkRouter(Ours)           & 51.5 & 24.0 & 44.0 & 42.0 & 15.8 & 15.9 & 37.0 & 32.89 & -0.25$\downarrow$ \\
\midrule
\multicolumn{10}{l}{\textbf{LLaVA-1.5-13B}} \\
Full Attention & 47.5 & 31.5 & 51.0 & 51.5 & 14.8 & 15.2 & 45.0 & 36.64 & -- \\
FastV          & 48.5 & 32.5 & 51.5 & 52.5 & 15.2 & 14.8 & 45.0 & 37.14 & +0.50$\uparrow$ \\
LOOK-M         & 47.5 & 30.0 & 50.5 & 51.5 & 13.6 & 13.5 & 9.5  & 30.87 & -5.77$\downarrow$ \\
SinkRouter(Ours)           & 47.5 & 31.5 & 51.0 & 51.5 & 15.4 & 16.6 & 44.0 & 36.79 & +0.15$\uparrow$ \\
\bottomrule
\end{tabular}
}
\end{table*}

Table~\ref{tab:llava-combined-transposed} further supports this pattern on the more challenging multimodal reasoning tasks in MileBench. SinkRouter remains close to Full Attention on both LLaVA backbones, indicating that the proposed decode-time routing transfers smoothly to multimodal reasoning workloads. LOOK-M shows noticeably  degradation on the benchmark average. FastV is competitive on the benchmark average and in some settings is slightly above the Full Attention baseline. However, this advantage is not uniformly reflected in all models and tasks. A more cautious reading of the results is that SinkRouter preserves multimodal performance at a level close to Full Attention, while avoiding the larger degradations seen in more aggressive compression-based baselines.

\subsection{Efficiency Evaluation}

\begin{figure}[tbp]
    \centering
    \includegraphics[width=\linewidth]{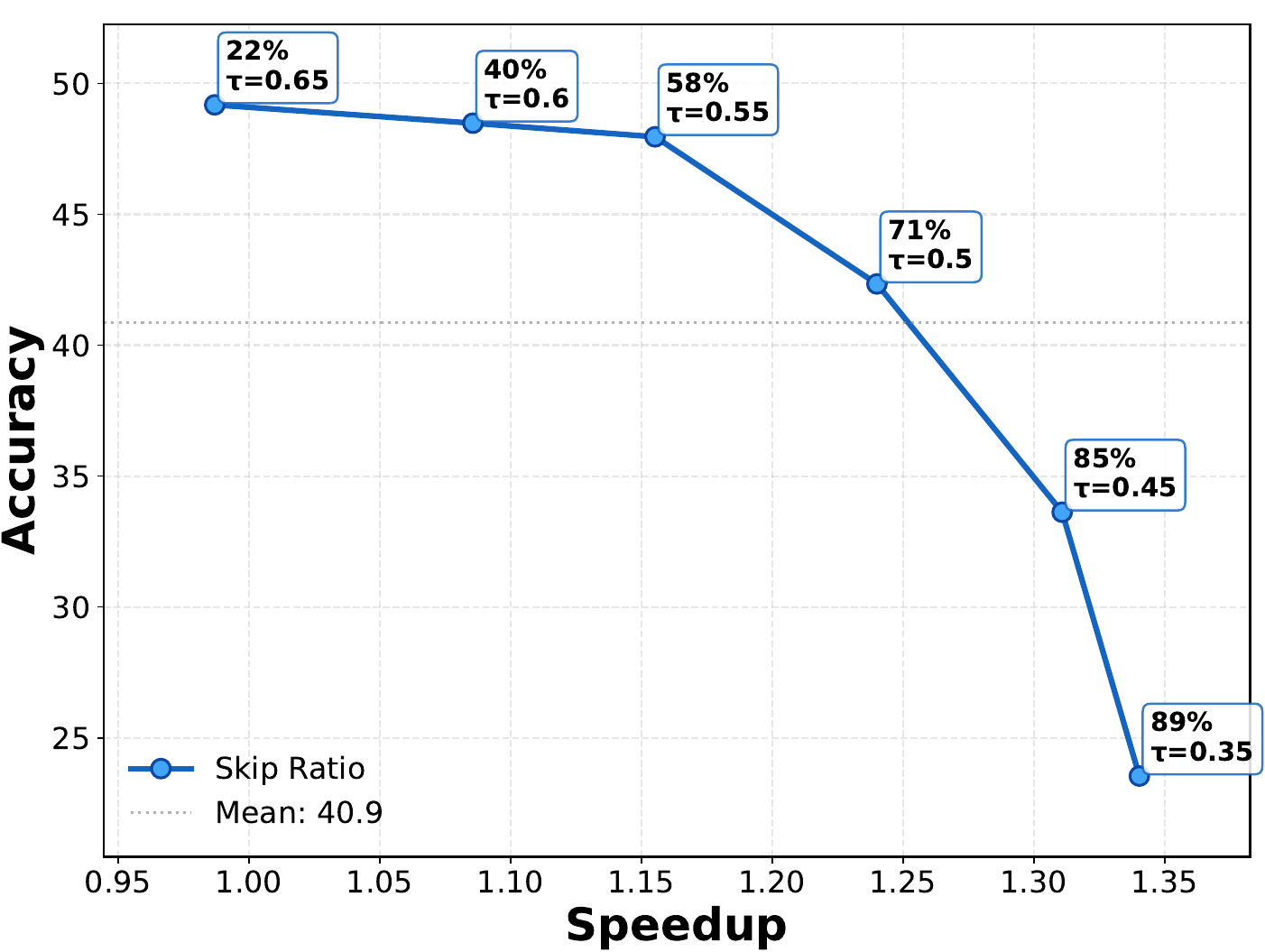}
    \caption{Accuracy--speed trade-off of SinkRouter at a fixed 128K context. Accuracy is measured by the average score on \textsc{LongBench}. Each point corresponds to a routing threshold and is annotated with the realized skip ratio.}
    \Description{A scatter plot showing the trade-off between speedup and accuracy for SinkRouter at a fixed 128K context. The x-axis is speedup and the y-axis is the average LongBench score. Each point corresponds to a routing threshold and is labeled with the realized skip ratio. The plot shows a mild trade-off before the point corresponding to tau equals 0.55 and about 58 percent skip, followed by a much steeper accuracy decline at larger skip ratios.}
    \label{fig:tradeoff}
\end{figure}

We evaluate the decoding efficiency of SinkRouter on a NVIDIA RTX PRO 6000 GPU. 
We first examine the accuracy--speed trade-off at a fixed 128K context. 
In Figure~\ref{fig:tradeoff}, each point corresponds to a routing threshold and is annotated with the realized skip ratio. 
The trade-off curve exhibits a clear transition around $\tau=0.55$. 
Before this point, increasing the skip ratio yields efficiency gains while the accuracy changes slightly.
Beyond this point, accuracy declines more rapidly as the skip ratio increases.
This behavior suggests that moderate skip ratios capture most of the efficiency gain while maintaining accuracy near the dense baseline.

\begin{figure}[tbp]
    \centering
    \includegraphics[width=\linewidth]{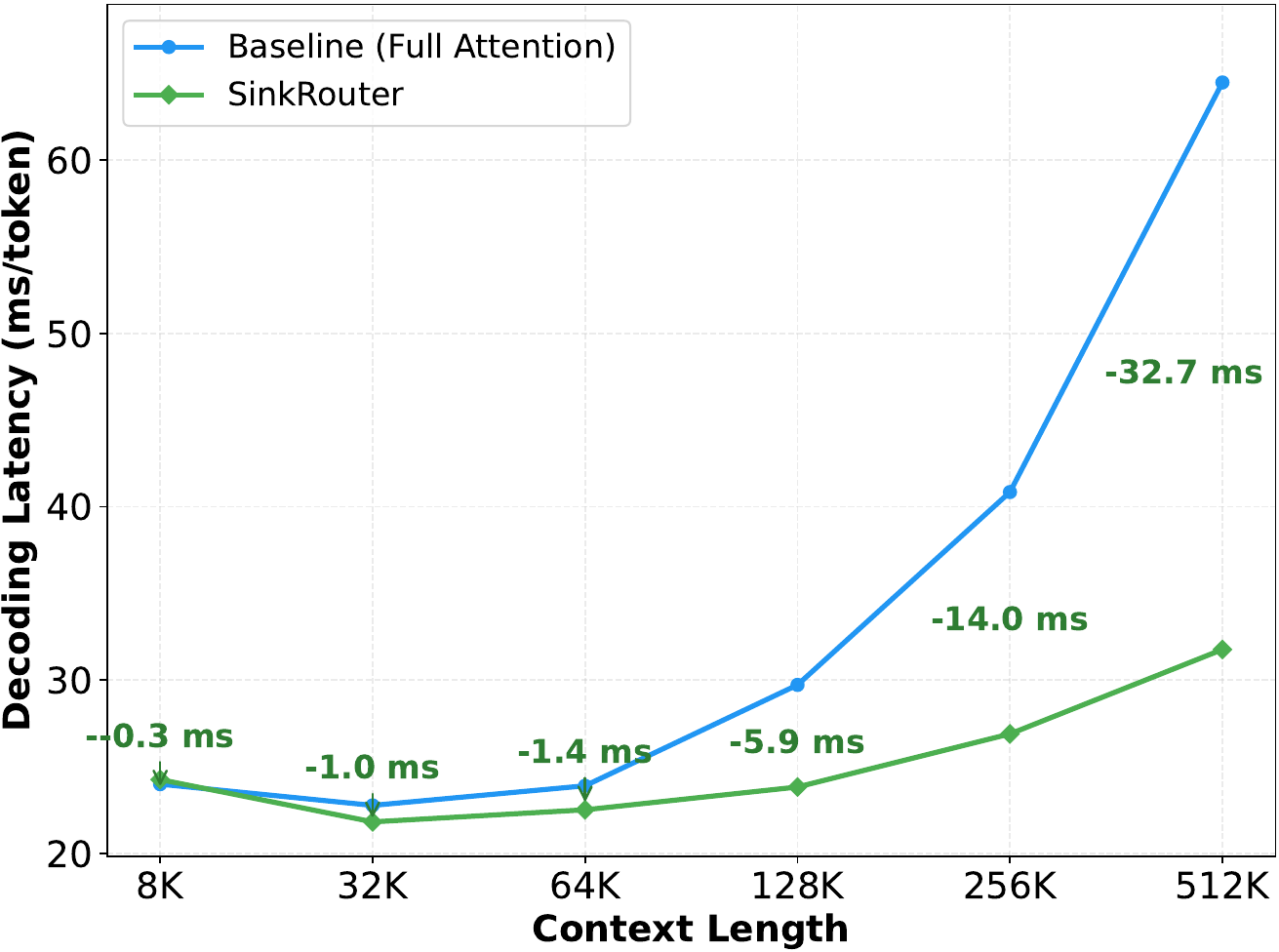}
    \caption{End-to-end per-token decoding latency across context lengths. SinkRouter consistently reduces decoding latency relative to the Full Attention baseline, and the absolute latency reduction becomes increasingly pronounced as the context grows.}
    \Description{A line plot comparing per-token decoding latency of Full Attention and SinkRouter across context lengths from 8K to 512K. SinkRouter remains below the Full Attention baseline at all lengths, and the absolute latency gap widens substantially at longer contexts.}
    \label{fig:latency_curve}
\end{figure}

We next evaluate end-to-end decoding latency across context lengths and SinkRouter uses a length-dependent threshold calibrated to achieve an average skip ratio of approximately 60\%. 
As shown in Figure~\ref{fig:latency_curve}, SinkRouter consistently reduces per-token decoding latency, and the absolute latency gap widens steadily as the context grows. 

\begin{figure}[tbp]
    \centering
    \includegraphics[width=\linewidth]{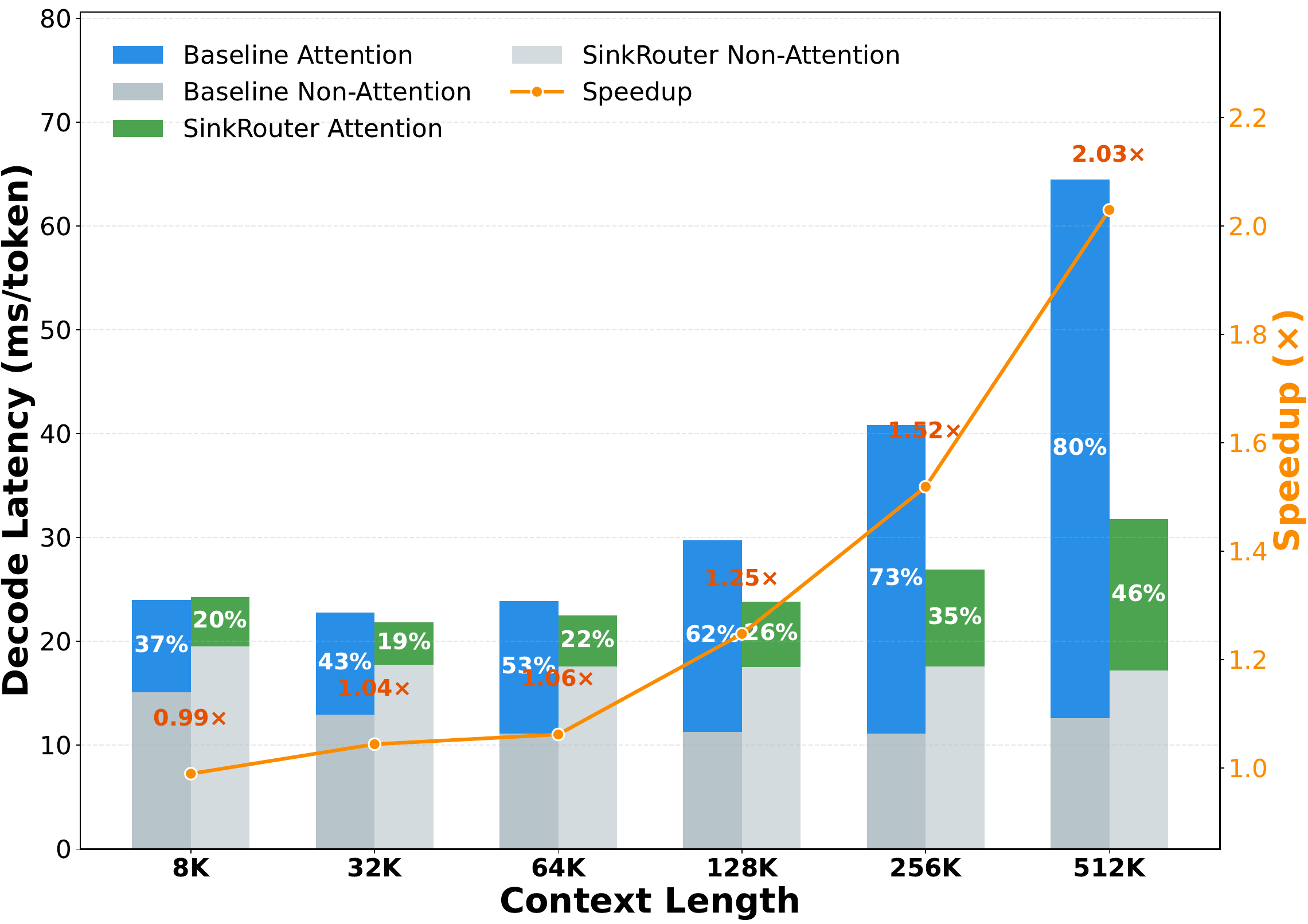}
    \caption{Latency composition and end-to-end speedup across context lengths. The stacked bars show the attention and non-attention contributions to per-token decoding latency for Full Attention and SinkRouter, while the line shows the resulting speedup.}
    \Description{A stacked bar chart showing the attention and non-attention components of per-token decoding latency for Full Attention and SinkRouter across context lengths from 8K to 512K, together with a line showing end-to-end speedup. The figure shows that the attention component grows with context length and that SinkRouter achieves larger speedups when the attention path becomes more dominant.}
    \label{fig:latency_breakdown}
\end{figure}

Figure~\ref{fig:latency_breakdown} further explains this behavior through latency composition. 
As the context grows, the attention path occupies an increasingly larger fraction of total decoding latency, while the non-attention path changes more modestly. 
Similarly, the end-to-end speedup of SinkRouter also increases with context length. 
These results show that both the absolute time saved and the resulting speedup become more pronounced as decoding is increasingly dominated by the attention path.

The gain remains limited before 64K. 
In this regime, historical KV-cache loading is still relatively modest, so memory-bandwidth pressure has not yet become the dominant bottleneck. 
Consequly, the benefit of SinkRouter mainly comes from its hardware-aware execution path, which avoids redundant memory traffic and tensor expansion associated with explicit \texttt{repeat\_kv} operations in grouped-query attention. 
As the context grows further, attention-related data movement accounts for a much larger share of total latency, so skipping routed groups could translate into end-to-end gains.

\section{Conclusion}

In this work, we present SinkRouter, a training-free routing framework that repurposes attention sinks as an emergent idling signal. By characterizing the initial token as a numerical vacuum with a near-zero Value norm, we enable the model to identify and bypass non-contributing attention groups during decoding. To achieve real-world efficiency, we developed a hardware-aware Triton kernel.Evaluated on diverse text and multimodal benchmarks, SinkRouter maintains near-lossless performance across Llama-3.1 and LLaVA-1.5 backbones, and achieves up to 2.03$\times$ faster decoding with a 512K context.

\bibliographystyle{ACM-Reference-Format}
\bibliography{acmart}

\clearpage
\appendix
\section*{Appendix}
\section{Threshold Calibration and Budget Stability}

We further analyze the calibration behavior of SinkRouter and the stability of its realized compute budget. Our goal is to understand whether a routing threshold can provide a predictable skip ratio across different tasks and context lengths, and how this behavior should be incorporated into the dynamic threshold used at inference time.

Our analysis proceeds in two stages. We first perform a threshold sweep at 16K context using 50 calibration samples, and measure the realized head skip ratio under different cosine-similarity thresholds. Based on this sweep, we select the threshold that gives a skip ratio closest to the target budget of 60\%, which yields $\tau=0.55$ in our setup. We then use this fixed threshold to evaluate stability across tasks and context lengths.

\begin{figure}[htbp]
    \centering
    \includegraphics[width=\linewidth]{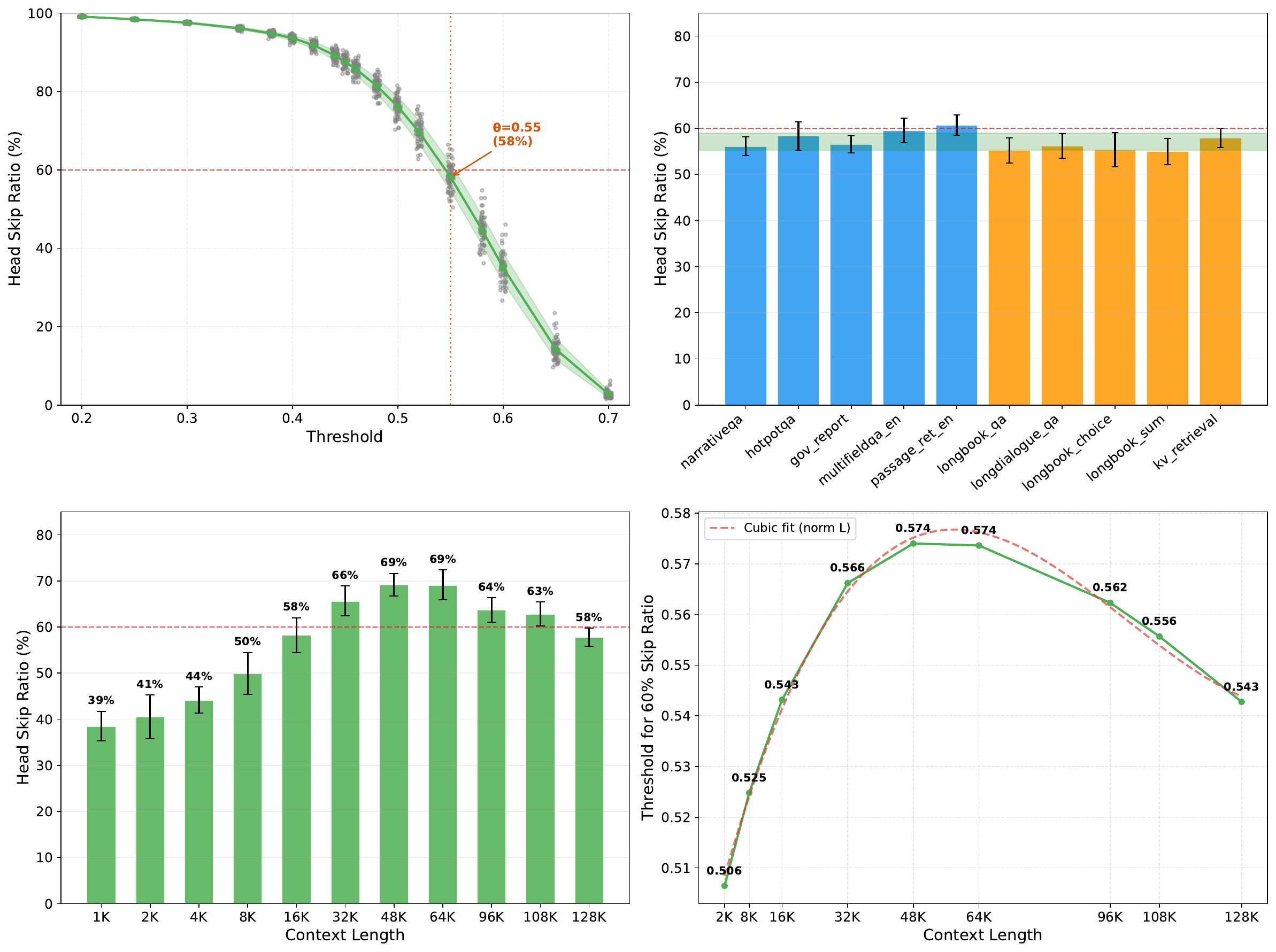}
    \caption{
    Threshold calibration, stability, and dynamic-threshold fitting.
    Top-left: threshold sweep at 16K context using 50 calibration samples;
    Top-right: realized skip ratios across \textsc{LongBench} and \textsc{InfiniteBench} tasks under the fixed threshold.
    Bottom-left: realized skip ratios across context lengths under the same fixed threshold.
    Bottom-right: threshold required to realize a 60\% skip ratio at different context lengths, together with a cubic fit over normalized context length.
    }
    \Description{
    A 2-by-2 figure showing threshold calibration and stability analysis for SinkRouter.
    Top-left: realized head skip ratio as a function of threshold at 16K context using 50 calibration samples, with tau equals 0.55 marked as approximately 58 percent skip.
    Top-right: realized skip ratios across LongBench and InfiniteBench tasks under the fixed threshold tau equals 0.55, showing limited variation across tasks.
    Bottom-left: realized skip ratios across context lengths under the same fixed threshold, showing a clear dependence on context length.
    Bottom-right: the threshold required to realize a 60 percent skip ratio at different context lengths, together with a cubic fit over normalized context length.
    Overall, the figure shows that the threshold is relatively stable across tasks but varies systematically with context length.
    }
    \label{fig:supp_stability}
\end{figure}

Figure~\ref{fig:supp_stability} summarizes three observations and one fitted calibration rule. First, the realized head skip ratio varies smoothly and monotonically with the threshold, which makes budget calibration well-defined. Second, when the calibrated threshold $\tau=0.55$ is applied across tasks from \textsc{LongBench} and \textsc{InfiniteBench}, the realized skip ratio shows only limited variation, indicating that task identity has relatively little influence on the operating point. Third, when the same threshold is applied across context lengths, the realized skip ratio changes systematically with length: it increases from short contexts to medium-length contexts, peaks around the mid-range, and then decreases again at longer contexts. This pattern shows that a single fixed threshold is not sufficient across a wide length range, even though it is reasonably stable across tasks.

Motivated by this result, we further calibrate the threshold required to realize a fixed 60\% skip ratio at different context lengths. The resulting threshold curve is itself non-monotonic: it first increases with context length, reaches its maximum in the mid-range, and then gradually decreases at longer contexts. To capture this behavior, we fit a cubic function of normalized context length,
\[
\tau(x)=ax^3+bx^2+cx+d,
\]
where $x$ denotes the normalized context-length variable. In practice, this fitted curve serves as the dynamic threshold function used by SinkRouter during decoding.

Taken together, these results suggest that the routing threshold is relatively stable across tasks but varies systematically with context length. This supports the use of lightweight calibration together with a length-dependent threshold function, rather than a single global threshold.

\section{Layer-Wise Residual-Stream Analysis of BOS-Dominant Heads}
\begin{figure*}[htbp]
    \centering
    \includegraphics[width=\linewidth]{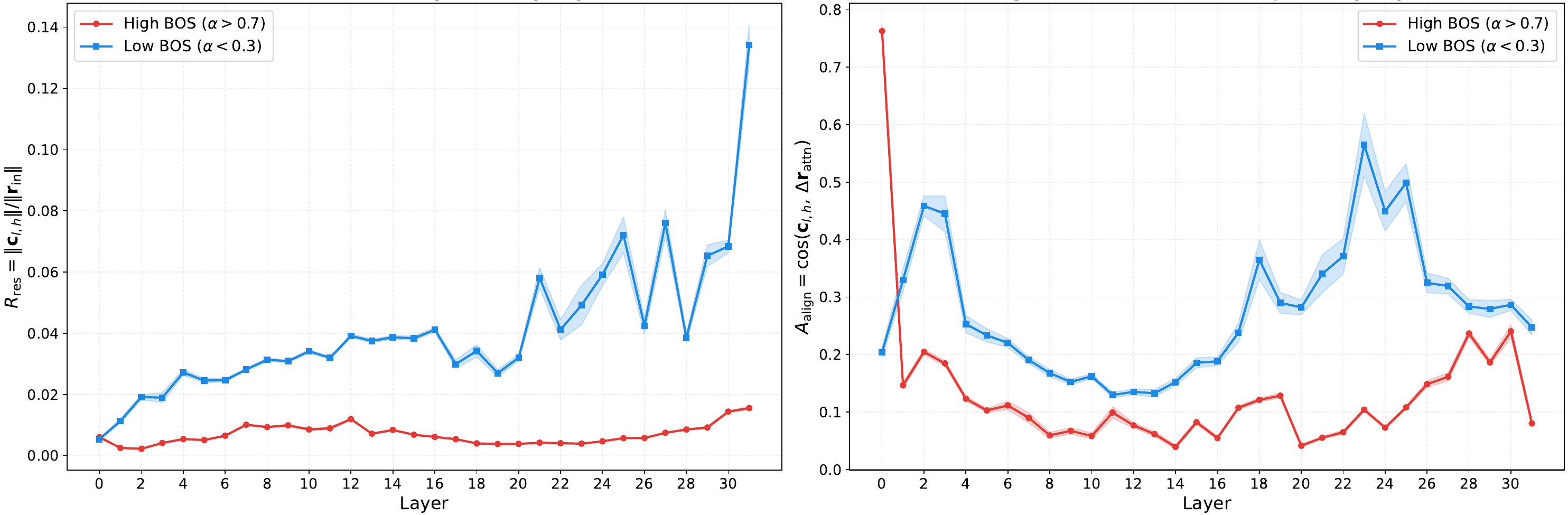}
    \caption{
    Layer-wise comparison between high-BOS heads and low-BOS heads.
    Left: residual-stream write magnitude across layers.
    Right: alignment with the aggregate attention update across layers.
    }
    \Description{
    Two plots comparing high-BOS heads and low-BOS heads across layers.
    The left plot shows that high-BOS heads have smaller residual-stream write magnitudes.
    The right plot shows that high-BOS heads are generally less aligned with the aggregate attention update direction.
    Together, the plots indicate that BOS-dominant heads contribute more weakly to the residual stream.
    }
    \label{fig:supp_residual_align}
\end{figure*}

To better understand whether BOS-dominant heads behave as weakly contributing computation, we analyze their effects on the residual stream at the layer level.
For each decoding step $t$, layer $l$, and head $h$, we define the vector written by that head into the residual stream as
\[
c_{l,h,t}=W_O^{(h)}o_{l,h,t},
\]
where $o_{l,h,t}$ denotes the head output before output projection and $W_O^{(h)}$ denotes the corresponding block of the output projection matrix.
We compare two groups of heads: high-BOS heads with $\alpha_{\mathrm{BOS}}>0.7$ and low-BOS heads with $\alpha_{\mathrm{BOS}}<0.3$.

In the left panel of Figure~\ref{fig:supp_residual_align}, the y-axis represents the residual write magnitude
\[
R_{\mathrm{res}}=\frac{\|c_{l,h,t}\|_2}{\|r^{\mathrm{in}}_{l,t}\|_2+\epsilon},
\]
which measures the magnitude of a head's residual-stream write relative to the input residual norm of the attention block.
Smaller values indicate weaker updates to the residual stream.

In the right panel, the y-axis represents the update-direction alignment
\[
A_{\mathrm{align}}=\cos\!\big(c_{l,h,t}, \Delta r^{\mathrm{attn}}_{l,t}\big),
\]
where
\[
\Delta r^{\mathrm{attn}}_{l,t}=r^{\mathrm{out}}_{l,t}-r^{\mathrm{in}}_{l,t}
\]
denotes the aggregate residual update produced by the attention block.
This quantity measures whether the vector written by a given head is aligned with the dominant update direction of the entire attention module.

As shown in Figure~\ref{fig:supp_residual_align}, high-BOS heads produce substantially smaller residual writes across most layers and are generally less aligned with the aggregate attention update than low-BOS heads. This behavior is consistent with the view that BOS-dominant routing corresponds to an evolved near-silent regime: while the model cannot literally avoid updating the residual stream, since the attention block must still return a valid output, the induced write is often weak enough to be functionally close to silent. We further observe that the first two layers behave atypically, particularly in alignment, suggesting a distinct role in early input processing. Accordingly, we exclude these two layers from routing in our actual implementation.

\section{Reliability of the Cosine Routing Proxy}
\label{sec:supp_proxy_reliability}

\begin{figure}[t]
    \centering
    \includegraphics[width=\linewidth]{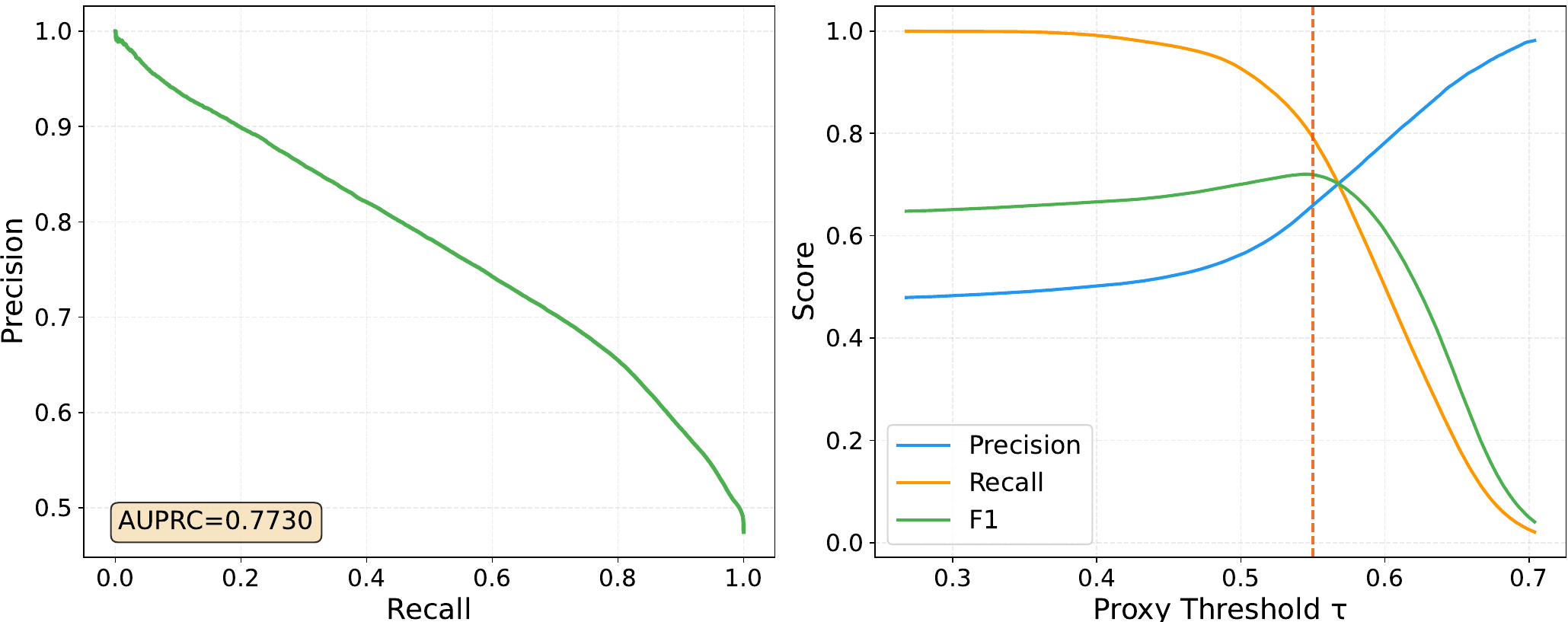}
    \caption{
    Predictive performance of $\cos(q, K_{\text{BOS}})$ as a routing proxy on \textsc{Llama-3.1-8B} at the KV-group level.
    Left: precision--recall curve for predicting the oracle sink event defined from full-attention BOS scores.
    Right: precision, recall, and F1 as functions of the proxy threshold $\tau$; the dashed vertical line marks the operating threshold $\tau=0.55$ used in SinkRouter.
    }
    \Description{
    Two plots evaluating the predictive performance of the cosine-similarity proxy on Llama-3.1-8B at the KV-group level.
    The left plot is a precision--recall curve with an area under the precision--recall curve of 0.7730.
    The right plot shows precision, recall, and F1 as functions of the proxy threshold tau, with a dashed vertical line at tau equals 0.55 indicating the operating threshold used in SinkRouter.
    }
    \label{fig:supp_proxy_pr}
\end{figure}

We further assess the reliability of $\cos(q, K_{\mathrm{BOS}})$ as the routing signal used by SinkRouter. Our goal is not only to determine whether the skipped set is enriched with sink-like heads, but also to quantify whether the proxy carries useful predictive information about the oracle sink event.

\begin{figure*}[t]
    \centering
    \includegraphics[width=\linewidth]{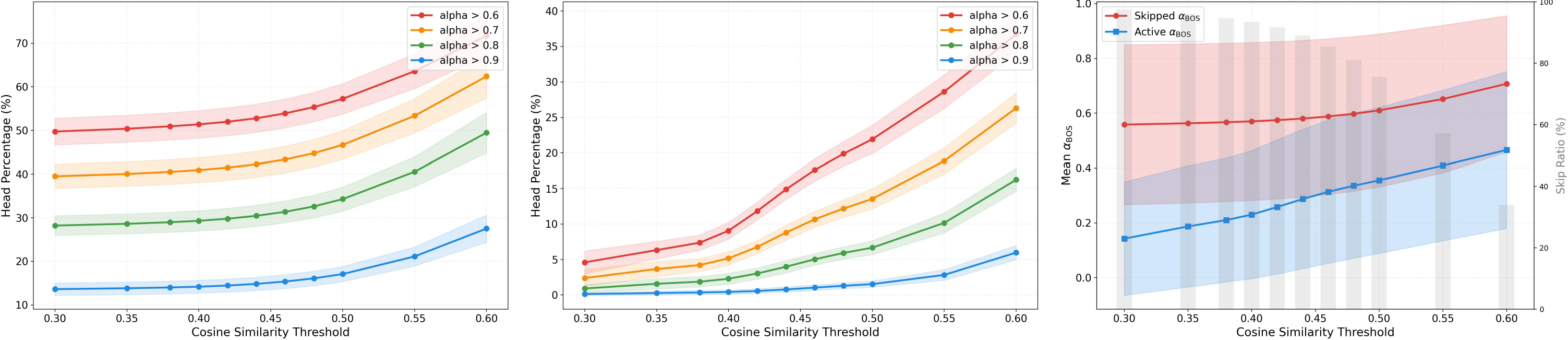}
    \caption{
    Distribution-level evidence for the cosine routing proxy at 16K context.
    Left: for skipped heads, the fraction whose BOS attention exceeds different cutoffs.
    Middle: the same statistics for active heads.
    Right: mean BOS attention of skipped and active heads as the cosine-similarity threshold varies; gray bars indicate the realized skip ratio.
    }
    \Description{
    Three plots showing the relationship between BOS attention and routing outcome at 16K context.
    The left plot shows, for skipped heads, the percentage whose BOS attention exceeds different cutoffs.
    The middle plot shows the same statistics for active heads.
    The right plot compares the mean BOS attention of skipped and active heads as the cosine-similarity threshold varies, with gray bars indicating the realized skip ratio.
    Across thresholds, skipped heads consistently appear more BOS-dominant than active heads.
    }
    \label{fig:supp_cosine_proxy}
\end{figure*}

We first examine this question from a distributional perspective at 16K context. Figure~\ref{fig:supp_cosine_proxy} compares BOS-attention statistics between skipped and active heads under a sweep of the cosine threshold. The left panel reports, for skipped heads, the fraction whose BOS attention exceeds a range of cutoffs, while the middle panel reports the same statistics for active heads. Across all tested BOS-attention cutoffs, skipped heads consistently contain a larger fraction of high-BOS heads than active heads. In particular, under the cutoff $\alpha_{\mathrm{BOS}}>0.6$, the skipped set remains dominated by high-BOS heads throughout the entire threshold range. The right panel further shows that the mean BOS attention of skipped heads remains consistently higher than that of active heads, although the variance within each set is non-negligible. This variance reflects heterogeneity across individual heads rather than a lack of separation: the enrichment statistics in the left and middle panels show that, at the population level, skipped heads remain consistently more BOS-dominant. Part of this variance is also expected because SinkRouter operates at the KV-group level rather than the individual-head level. Since each group aggregates multiple query heads, both proxy scores and oracle labels are necessarily coarser and may mix heads with different degrees of BOS dominance. This makes the separation less sharp than in a purely head-level analysis, while keeping the evaluation aligned with the actual routing granularity used by SinkRouter.

We next quantify the predictive value of the same proxy at the actual routing granularity of SinkRouter, i.e., the KV-group level, using \textsc{Llama-3.1-8B}. We cast sink detection as a binary classification problem. At the KV-group level, we define the oracle sink label using the full-attention BOS score of the group, and use the corresponding group-level cosine score as the proxy prediction score. Figure~\ref{fig:supp_proxy_pr} shows that this proxy achieves useful predictive performance, with an AUPRC of 0.7730. The right panel further illustrates the precision--recall trade-off induced by the proxy threshold. As the threshold increases, precision improves steadily while recall decreases, and the operating threshold used in SinkRouter, $\tau=0.55$, lies in a favorable region where F1 remains close to its peak. This suggests that the selected threshold is not only compatible with the target skip budget, but also achieves a favorable balance between proxy reliability and sink coverage.

Taken together, Figures~\ref{fig:supp_cosine_proxy} and~\ref{fig:supp_proxy_pr} provide complementary evidence for the reliability of $\cos(q, K_{\mathrm{BOS}})$. The former shows that the skipped set is qualitatively enriched with BOS-dominant heads, whereas the latter shows that the proxy has quantitatively useful predictive value for the oracle sink event at the same KV-group granularity used by SinkRouter. These results support $\cos(q, K_{\mathrm{BOS}})$ as a reliable and lightweight pre-attention routing signal.

\end{document}